\documentclass[runningheads]{llncs}

\usepackage{eccv}

\usepackage{eccvabbrv}
\usepackage{multirow}
\usepackage{algorithm}
\usepackage{algorithmic}
\usepackage{xcolor}
\usepackage{caption,graphicx,wrapfig}

\usepackage{graphicx}
\usepackage{booktabs}

\definecolor{shapecolor}{rgb}{0.0,0.5,0.0}

\usepackage[accsupp]{axessibility}  %

\usepackage[hidelinks]{hyperref}

\usepackage{orcidlink}

\usepackage[misc,geometry]{ifsym}

\usepackage{enumitem}
\setitemize{noitemsep,topsep=0pt,parsep=0pt,partopsep=0pt}
\newcommand\extrafootertext[1]{%
    \bgroup
    \renewcommand\thefootnote{\fnsymbol{footnote}}%
    \renewcommand\thempfootnote{\fnsymbol{mpfootnote}}%
    \footnotetext[0]{#1}%
    \egroup
}

\usepackage{float}
\floatstyle{plaintop}
\restylefloat{table}
\usepackage{graphicx}
\usepackage[dvipsnames]{xcolor}
\usepackage[title]{appendix}%

\begin{document}

\title{
Motion Mamba: Efficient and Long Sequence Motion Generation 
% with Hierarchical and Bidirectional Selective SSM
}

\titlerunning{Motion Mamba}

\author{Zeyu Zhang\inst{1}\inst{2}$^{*}$$^{\dagger}$\orcidlink{0009-0006-8819-3741} \and
Akide Liu\inst{1}$^{*}$\orcidlink{0000-0001-9870-8303}
\and Ian Reid\inst{3}\orcidlink{0000-0001-7790-6423}
 \and Richard Hartley\inst{2}\orcidlink{0000-0002-5005-0191} \and Bohan Zhuang\inst{1}\orcidlink{0000-0002-0074-0303} \and
Hao Tang\inst{4}$^{\text{\Letter}}$\orcidlink{0000-0002-2077-1246}}

\authorrunning{Zeyu and Akide et al.}

\institute{
Monash University \and
The Australian National University \and
Mohamed bin Zayed University of Artificial Intelligence \and
National Key Laboratory for Multimedia Information Processing, \\
School of Computer Science, Peking University}

\extrafootertext{\fontsize{7}{4}\selectfont $^{*}$Equal contribution. \\
\fontsize{7}{4}\selectfont $^{\dagger}$Work done while being a research assistant at Monash University.\\
\fontsize{7}{4}\selectfont $^{\text{\Letter}}$Corresponding author: \href{bjdxtanghao@gmail.com}{bjdxtanghao@gmail.com}}

\maketitle

\begin{center}
\vspace{-0.6cm}
  \href{https://steve-zeyu-zhang.github.io/MotionMamba}{\textcolor{blue}{\textbf{\fontsize{7}{4}\selectfont https://steve-zeyu-zhang.github.io/MotionMamba}}}
\end{center}

\begin{center}
 \centering
 \captionsetup{type=figure}
 \includegraphics[width=1\textwidth]{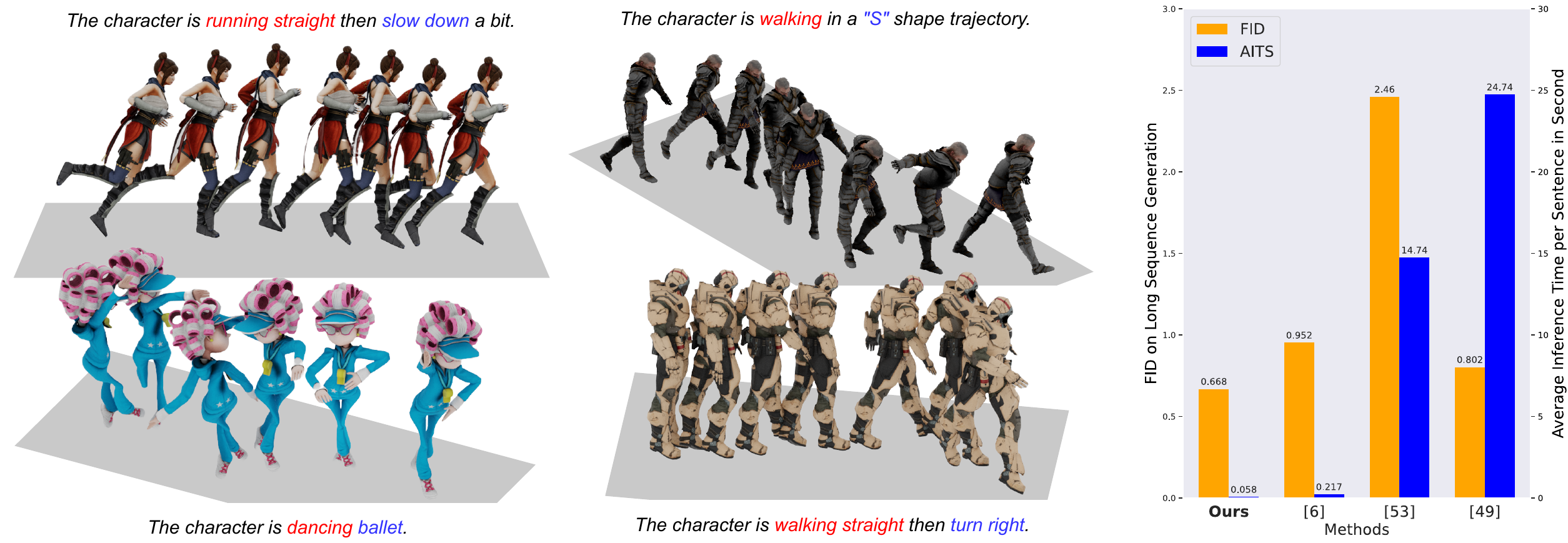}
 \caption{\textbf{Motion Mamba} has achieved significantly superior performance on long squence modeling and motion generation efficiency compared with other well-designed state-of-the-art methods such as MLD \cite{chen2023executing}, MotionDiffuse \cite{zhang2024motiondiffuse}, and MDM \cite{tevet2022human}.}
\vspace{-0.3cm}
  \label{fig:demo}
\end{center}

\begin{abstract}

Human motion generation stands as a significant pursuit in generative computer vision,
while achieving long-sequence and efficient motion generation remains challenging.
Recent advancements in state space models (SSMs), notably Mamba, have showcased considerable promise in long sequence modeling with an efficient hardware-aware design, which appears to be a significant direction upon building motion generation model.
Nevertheless, adapting SSMs to motion generation faces hurdles since the lack of a specialized design architecture to model motion sequence.
To address these challenges, we 
propose \textbf{Motion Mamba}, a simple yet efficient approach that presents the pioneering motion generation model
utilized SSMs. 
Specifically, we design
a \textbf{Hierarchical Temporal Mamba} (\textbf{HTM}) block to process temporal data by 
ensemble varying numbers of isolated SSM modules across a symmetric U-Net architecture 
aimed at preserving motion consistency between frames.
We also design
a \textbf{Bidirectional Spatial Mamba} (\textbf{BSM}) block to bidirectionally process latent poses, 
to enhance accurate motion generation within a temporal frame.
Our proposed method achieves up to \textbf{50\%} FID improvement and up to \textbf{4} times faster on the HumanML3D and KIT-ML datasets compared to the previous best diffusion-based method, which demonstrates strong capabilities of high-quality long sequence motion modeling and real-time human motion generation.

\end{abstract}

\section{Introduction}
\label{sec:intro}

% Human motion generation stands as a holy grail in generative computer vision, holding broad applications in computer animation, game development, and robot manipulation. 
Human motion generation involves creating high-quality 3D human motion based on given conditions, such as a text prompt, which generates Euler angles for each joint in a human skeleton.
To emulate human motion effectively, virtual characters must respond to the conditional context, exhibit natural movement, and perform motion accurately. 
Recent motion generation models are categorized into four main approaches: autoencoder-based \cite{ahuja2019language2pose, tevet2022motionclip, petrovich2022temos, guo2022generating, zhong2023attt2m, gong2023tm2d}, utilizing transformers for latent space compression and motion synthesis; GAN-based \cite{lin2018human, harvey2020robust, barsoum2018hp}, using discriminators to enhance the realism of generated motions; autoregressive models \cite{jiang2024motiongpt}, treating motion sequences as languages with specialized codebooks; and diffusion-based \cite{zhang2024motiondiffuse, tevet2022human, chen2023executing}, employing denoising steps for motion generation. 
Challenges vary across methods, with autoencoder models struggling to generate accurate motions from detailed descriptions due to textual information compression, GAN-based models facing training difficulties, especially in conditional tasks, and diffusion-based models relying on complex transformer-based architectures results in inefficient of motion prediction.

Although diffusion-based models excel at generating motion with robust performance and often exhibit superior diversity, they encounter two limitations. 
1) Convolutional or transformer-based diffusion methods exhibit limitations in generating \textit{long-range motion sequences}. Previous transformer-based methodologies \cite{zhang2024motiondiffuse, tevet2022human, chen2023executing} have focused on modeling long-range dependencies and acquiring comprehensive global context information. Despite these advances, they are frequently associated with a substantial increase in computational requirements. Furthermore, transformer architectures are not intrinsically designed for temporal sequential modeling, which poses an inherent limitation.
2) The \textit{efficiency} of inference in transformer-based diffusion methods is constrained. Although prior research has attempted to leverage the Variational Autoencoder for denoising operations in the latent space \cite{chen2023executing}, the inference speed remains adversely affected by the attention mechanism's quadratic scaling, leading to inefficient motion generation.
Consequently, exploring a new architectural paradigm that accommodates long-range dependencies and maintains a linear computational complexity is crucial for sustaining motion generation tasks.

Recent advances have sparked renewed interest in state space models (SSMs) \cite{gu2021efficiently, gu2021combining}, a field that originated from the foundational classic state space model \cite{kalman1960new}.  
Modern versions of SSMs stand out due to their ability to effectively capture long-range dependencies, a capability greatly improved by the introduction of parallel training techniques.
This evolution has led to various methodologies based on SSM, notably the linear state space layers (LSSL) \cite{gu2021combining}, the structured state-space sequence model (S4) \cite{gu2021efficiently}, the diagonal state space (DSS) \cite{gupta2022diagonal}, and S4D \cite{gu2022parameterization}. 
These methods have been carefully designed to handle sequential data across various tasks and modalities, paying special attention to modeling long-range dependencies.
Their efficacy in managing long sequences is attributed to the implementation of convolutional computations \cite{gu2021combining} and near-linear computational strategies, such as mamba \cite{gu2023mamba}, marking a significant stride in sequentially oriented tasks, including large language model decoding and motion sequence generation.

Adapting selective state space modules for motion generation tasks presents notable challenges, primarily due to the lack of specialized design in SSMs for capturing the sensitive motion details
required for temporal representation and the complexities involved in aggregating latent space. In response to these challenges, we have meticulously developed a motion generation architecture, specifically tailored to address the intricacies of long-term sequence generation, while optimizing for computational efficiency with near-linear-time complexity. This innovation is embodied in the Motion Mamba model, a simple yet potent approach to motion generation.
The Motion Mamba framework pioneers a diffusion-based generative system, incorporating two key components oriented toward SSM as shown in Figure. \ref{fig:block}:
(1) a \textbf{H}ierarchical \textbf{T}emporal \textbf{M}amba (HTM) block: This component is ingeniously crafted to arrange motion frames in sequential order, using hierarchically adjusted scanning. It is adept at identifying temporal dependencies at various depths, thereby facilitating a thorough comprehension of the dynamics inherent in motion sequences.
(2) a \textbf{B}idirectional \textbf{S}patial \textbf{M}amba (BSM) block: This block is designed to unravel the structured latent skeleton by evaluating data from both forward and reverse directions. Its primary goal is to safeguard the continuity of information flow, significantly bolstering the model's capacity for precise motion generation through the retention of dense informational exchange.

The Motion Mamba introduces a new approach to motion generation that strikes an exceptional trade-off between accuracy and efficiency, shown in Fig.~\ref{fig:demo}. Our experimental results underscore the significant improvements brought about by Motion Mamba, showcasing a remarkable improvement in the Fréchet Inception Distance (FID), with a reduction of up to 50\% from the prior state-of-the-art metric of 0.473 to an impressive 0.281 on the HumanML3D dataset \cite{guo2022generating}. Furthermore, we emphasize our framework's unparalleled inference speed, which is four times faster than previous methods, achieving an average inference time of only 0.058 seconds per sequence compared to the 0.217 seconds required by the MLD \cite{chen2023executing} method per sequence. These outcomes unequivocally establish Motion Mamba's state-of-the-art performance, concurrently ensuring fast inference speeds for conditional human motion generation tasks.

Our contributions to the field of motion generation can be summarized as:
\vspace{-0.14cm}
\begin{enumerate}[leftmargin=*]
    \item We introduce a simple yet effective framework, named \textit{Motion Mamba}, which is a pioneering method integrates a selective scanning mechanism into motion generation tasks.
    \item \textit{Motion Mamba} is comprised of two modules: Hierarchical Temporal Mamba (HTM) and Bidirectional Spatial Mamba (BSM), which are designed for temporal and spatial modeling, respectively. HTM blocks are tasked with processing temporal motion data, aiming to enhance motion consistency across frames. BSM blocks are engineered to bidirectionally capture the channel-wise flow of hidden information within the latent pose representations.
    \item \textit{Motion Mamba} framework demonstrated exceptional performance on text-to-motion generation task, through experimental validation on the HumanML3D \cite{guo2022generating} and KIT-ML \cite{plappert2016kit} datasets. Our methodology achieved state-of-the-art generation quality and significantly improved long-squence modeling, meanwhile optimizing inference speed.

\end{enumerate}

\section{Related Works}

\noindent\textbf{Human Motion Generation.}
Generating human motion is a significant application of computer vision, essential for various applications like 3D modelling and robot manipulation. Recently, the predominant method of achieving human motion generation, known as the \textit{Text-to-Motion} task, involves learning a common latent space for both language and motion. 

DVGAN \cite{lin2018human} creates the GAN \cite{goodfellow2014generative} discriminator by densely validating at each time-scale and perturbing the discriminator input for translation invariance, enabling motion generation and completion. 
ERD-QV \cite{harvey2020robust} enhances latent representations through two additive modifiers: a time-to-arrival embedding applied universally and an additive scheduled target noise vector used during extended transitions. It further improves transition quality by incorporating a GAN framework with two discriminators operating at different timescales.
HP-GAN \cite{barsoum2018hp}, trained with a modified version of the improved WGAN-GP \cite{gulrajani2017improved}, utilizes a custom loss function designed for human motion prediction. It learns a probability density function of future human poses conditioned on previous poses.

Autoencoders \cite{rumelhart1985learning, kramer1991nonlinear} are notable generative models known for their ability to represent data robustly by compressing high-dimensional data into a latent space, which is widely adopted in human motion generation \cite{zhang2024motionavatar,zhang2024infinimotion}. JL2P \cite{ahuja2019language2pose} uses RNN-based autoencoders \cite{hopfield1982neural} to learn a combined representation of language and pose. It restricts the direct mapping from text to motion to a one-to-one relationship. MotionCLIP \cite{tevet2022motionclip} uses Transformer-based Autoencoders \cite{vaswani2017attention} to reconstruct motion while ensuring alignment with the corresponding text label in the CLIP \cite{radford2021learning} space. This alignment effectively integrates the semantic knowledge from CLIP into the human motion manifold. TEMOS \cite{petrovich2022temos} and T2M \cite{guo2022generating} combine a Transformer-based VAE \cite{kingma2014auto} with a text encoder to generate distribution parameters that work within the VAE latent space. AttT2M \cite{zhong2023attt2m} and TM2D \cite{gong2023tm2d} incorporate a body-part spatio-temporal encoder into VQ-VAE \cite{van2017neural} for enhanced learning of a discrete latent space with increased expressiveness. 

Diffusion models \cite{sohl2015deep, ho2020denoising, dhariwal2021diffusion, rombach2022high} have recently surpassed GANs and VAEs in generating 2D images. Developing a motion generation model based on diffusion models is obviously an attractive direction. MotionDiffuse \cite{zhang2024motiondiffuse} introduces the inaugural framework for text-driven motion generation based on diffusion models. It showcases several desirable properties, including probabilistic mapping, realistic synthesis, and multi-level manipulation. MDM \cite{tevet2022human} utilizes a classifier-free Transformer-based diffusion model for the human motion domain to predict sample rather than noise in each diffusion step. MLD \cite{chen2023executing} performs a diffusion process in latent motion space, rather than using a diffusion model to establish connections between raw motion sequences and conditional inputs.

\noindent\textbf{State Space Models.}
Recently, state space sequence models (SSMs) \cite{gu2021efficiently, gu2021combining}, drawing inspiration from classical state-space models \cite{kalman1960new}, have emerged as a promising architecture for sequence modeling \cite{hu2024zigma}. Mamba \cite{gu2023mamba} introduces a selective SSM architecture, integrating time-varying parameters into the SSM framework, and proposes a hardware-aware algorithm to facilitate highly efficient training and inference processes. Some research works leverage SSM in computer vision to process 2D data. The 2D SSM \cite{baron20232} introduces an SSM block at the beginning of each transformer block \cite{vaswani2017attention, dosovitskiy2020image}. This approach aims to achieve efficient parameterization, accelerated computation, and a suitable normalization scheme. SGConvNeXt \cite{li2022makes} presents a structured global convolution method inspired by ConvNeXt \cite{liu2022convnet}, incorporating multi-scale sub-kernels to achieve both parameterization efficiency and effective long sequence modeling. ConvSSM \cite{smith2024convolutional} integrates the tensor modeling principles of ConvLSTM \cite{shi2015convolutional} with SSMs, elucidating the utilization of parallel scans in convolutional recurrences. This approach enables subquadratic parallelization and rapid autoregressive generation. Vim \cite{zhu2024vision} introduces a bidirectional SSM block \cite{wang2022pretraining} for efficient and versatile visual representation learning, achieving performance comparable to established ViT \cite{dosovitskiy2020image} methods. VMamba \cite{liu2024vmamba} introduces a Cross-Scan Module (CSM) designed to traverse the spatial domain and transform any non-causal visual image into ordered patch sequences. This approach achieves linear complexity while preserving global receptive fields. There have also been attempts to utilize SSMs to handle higher-dimensional data. Mamba-ND \cite{li2024mamba} explores various combinations of SSM and different scan directions within the SSM block to adapt Mamba \cite{gu2023mamba} to higher-dimensional tasks. 
Recent efforts have sought to replace the traditional transformer-based U-Net within the diffusion denoiser with the SSM block, with the aim of enhancing image generation efficiency. DiffuSSM \cite{yan2023diffusion} adeptly manages higher resolutions without relying on global compression, thus maintaining detailed image representation throughout the diffusion process.

\section{The Proposed Method}

In this section, we delineate the architecture and operational principles of the \emph{Motion Mamba} framework, designed for generating human motion over long ranges efficiently from textual descriptions. Initially, we discuss the foundational concepts underpinning our approach, including the Mamba Model \cite{gu2023mamba} and the latent diffusion model \cite{chen2023executing}. Following this, we detail our uniquely crafted architecture that leverages the Mamba model to enhance motion generation efficiency. This architecture comprises two principal components: the Hierarchical Temporal Mamba (HTM) block, which addresses temporal aspects, and the Bidirectional Spatial Mamba (BSM) block, focusing on spatial dynamics.

\subsection{Preliminaries}

\noindent\textbf{Selective Structured State Space Sequence Model.} 
SSMs particularly through the contributions of structured state space sequence models (S4) and Mamba, have demonstrated exceptional proficiency in handling long sequences. These models operationalize the mapping of a 1-D function or sequence, $x(t) \in \mathbb{R} \mapsto y(t) \in \mathbb{R}$, through a hidden state $h(t) \in \mathbb{R}^N$, employing $\mathbf{A} \in \mathbb{R}^{N \times N}$ as the evolution parameters, $\mathbf{B} \in \mathbb{R}^{N \times 1}$ and $\mathbf{C} \in \mathbb{R}^{1 \times N}$ as the projection parameters, respectively.
% The continuous system dynamics are described by the ordinary differential equation (ODE):
% \begin{equation}
% \begin{aligned}
% h'(t) &= \mathbf{A}h(t) + \mathbf{B}x(t), \\
% y(t) &= \mathbf{C}h(t) + \mathbf{D}x(t),
% \end{aligned}
% \end{equation}
% with $x(t)$ representing a continuous input signal and $y(t)$ a continuous output signal in the time domain.

% To adapt these continuous dynamics for practical computation, the S4 and Mamba models employ a discretization process, notably using the zero-order hold (ZOH) method, resulting in a transformation of continuous parameters into discrete ones:
% \begin{equation}
% \begin{aligned}
% \mathbf{\overline{A}} &= \exp{(\mathbf{\Delta}\mathbf{A})}, \\
% \mathbf{\overline{B}} &= (\mathbf{\Delta} \mathbf{A})^{-1}(\exp{(\mathbf{\Delta} \mathbf{A})} - \mathbf{I}) \cdot \mathbf{\Delta} \mathbf{B}.
% \end{aligned}
% \end{equation}
The discretized system can then be expressed as follows, incorporating a step size $\mathbf{\Delta}$:
\begin{equation}
\begin{aligned}
h_t &= \mathbf{\overline{A}}h_{t-1} + \mathbf{\overline{B}}x_{t}, \\
y_t &= \mathbf{C}h_t.
\end{aligned}
\end{equation}
This adaptation facilitates the computation of output through global convolution, leveraging a structured convolutional kernel $\overline{\mathbf{K}}$, which encompasses the entire length $M$ of the input sequence $\mathbf{x}$:
\begin{equation}
\begin{aligned}
\mathbf{\overline{K}} &= (\mathbf{C}\mathbf{\overline{B}}, \mathbf{C}\mathbf{\overline{A}}\mathbf{\overline{B}}, \dots, \mathbf{C}\mathbf{\overline{A}}^{M-1}\mathbf{\overline{B}}), \\
\mathbf{y} &= \mathbf{x} * \mathbf{\overline{K}}.
\end{aligned}
\end{equation}
Selective models like Mamba introduce time-varying parameters, deviating from the linear time invariance (LTI) assumption and complicating parallel computation. However, hardware-aware optimizations, such as associative scans, have been developed to address these computational challenges, highlighting the ongoing evolution and application of SSMs in modeling complex temporal dynamics.

\noindent\textbf{Latent Motion Diffusion Model.}
Diffusion probabilistic models offer a significant advancement in motion generation by gradually reducing noise from a Gaussian distribution to a target data distribution $p(x)$ through a T-length learned Markov process ~\cite{sohl2015deep, dhariwal2021diffusion, ho2020denoising, saharia2022image, rombach2022high, tevet2022human, zhang2024motiondiffuse} , giving $\{\boldsymbol{x_t}\}^{T}_{t=1}$. In the motion generation, we define our trainable diffusion models with a denoiser $\epsilon_\theta\left(x_t, t\right)$ which anneal the random noise to motion sequence $\{{\hat{x}_t^{1:N}}\}^{T}_{t=1}$ iteratively.
To address the inefficiencies of applying diffusion models directly to raw motion sequences, we employ a low-dimensional motion latent space for the diffusion process.
Given an input condition c, such as a descriptive sentence $\boldsymbol{w}^{1:N}=\{w^i\}_{i=1}^{N}$, an action label $a$ from a predefined set $\mathcal{A}$, or an empty condition $c = \varnothing$, and the motion representation that combines 3D joint rotations, positions, velocities, and foot contact as proposed in~\cite{guo2022generating}.
The frozen CLIP \cite{radford2021learning} text encoder  $\tau_\theta^{w}$ has been employed to obtain projected text embedding $\tau_\theta^w(w^{1:N}) \in \mathbb{R}^{1 \times d}$, thereby conditional denoiser comprised in term of $\epsilon_\theta(z_t,t,\tau_\theta(c))$.
The latent diffusion model $\epsilon_\theta\left(x_t, t\right)$ aimed to generate the human motion sequence in terms of  $\hat{x}^{1:L}=\{\hat{x}^i\}_{i=1}^{L}$, where L denotes the sequence length or number of frames~\cite{petrovich22temos, petrovich21actor, vposer_SMPL-X:2019, zhang2021we}. Afterthat we reused the motion Variational AutoEncoder (VAE) $\mathcal{V}=\{\mathcal{E}, \mathcal{D}\}$ proposed in MLD \cite{chen2023executing} to manipulate the motion sequence in latent space $z = \mathcal{E}(x^{1:L})$, and decompress the intermediate representation to motion sequence by $\hat{x}^{1:L} = \mathcal{D}(z) = \mathcal{D}{\mathcal{E}(x^{1:L})}$ \cite{vaswani2017attention, ronneberger2015u, kingma2014auto}.
Finally, our latent diffusion model is trained with an objective focusing on minimization of MSE between true and predicted noise in the latent space, facilitating efficient and high-quality motion generation~\cite{bao2022all, ho2020denoising}.

\begin{figure}[t]
    \centering
    \includegraphics[width=1\linewidth]{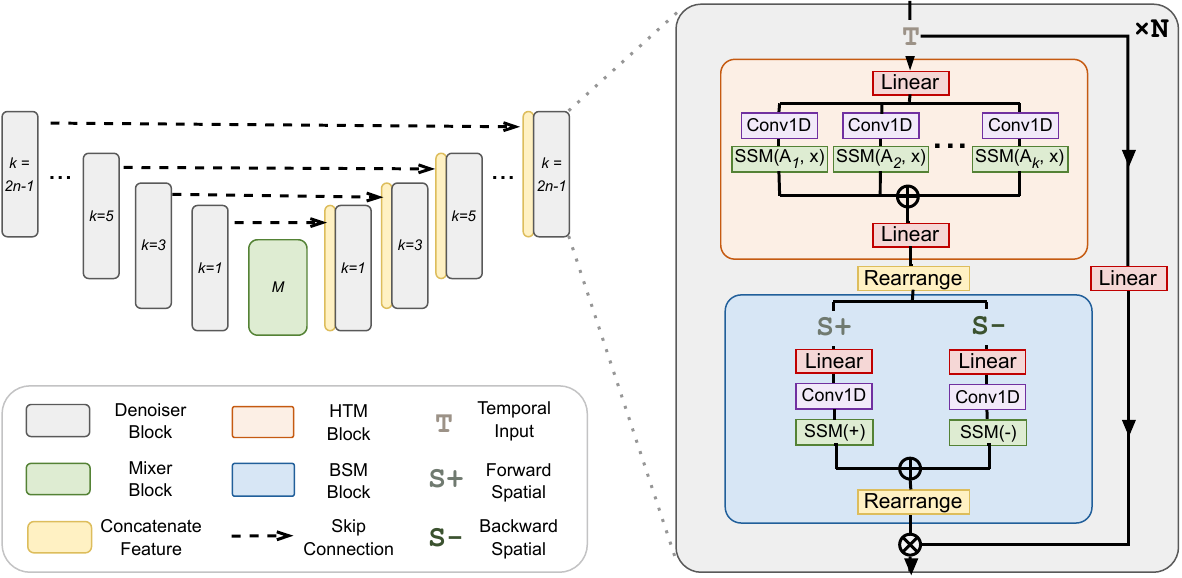} 
    \caption{This figure illustrates the architecture of the proposed Motion Mamba model. Each of encoder and decoder blocks consists of a Hierarchical Temporal Mamba block (HTM) and a Bidirectional Spatial Mamba (BSM) block, which possess hierarchical scan and bidirectional scan within SSM layers respectively. This symmetric distribution of scans ensure a balanced and coherence framework across the encoder-decoder architecture.}
    \label{fig:block}
    \vspace{-0.4cm}
\end{figure}

\subsection{Motion Mamba}

The architecture of the proposed \emph{Motion Mamba} framework is illustrated in Figure. \ref{fig:block}. At its core, Motion Mamba utilizes a denoising U-Net architecture, which is distinguished for its effectiveness in modeling the continuous, temporal sequences of motion frames. This effectiveness is attributed to the inherent long-sequence modeling capacity of the Mamba model. The denoiser, denoted by $\epsilon_\theta$, comprises $N$ blocks including encoder $E_{1..N}$ and decoder $D_{1..N}$ . Additionally, the architecture is enhanced with a transformer-based attention mixer block $M$, designed to augment the model's ability to capture complex temporal dynamics.
\begin{equation}
\epsilon_\theta(x) \equiv \{E_{1...N},M,D_{1..N}\}.
\end{equation}

\begin{algorithm}[t]
\caption{Hierarchical Temporal Mamba (HTM) Block.}
\label{algo:htm}
\small
\begin{algorithmic}[1]
\REQUIRE{compressed latent representations $z$ : $(T, B, C)$}
\ENSURE{transformed representations $z_{\text{HTM}}$ : $(T, B, E)$}
\STATE \textcolor{gray}{\text{/* linear projection layer */}}
\STATE $x, z$ : $(T, B, E)$ $\leftarrow$ Linear$(z)$
\STATE \textcolor{gray}{\text{/* set of scans and memory matrices */}}
\STATE $K=\{S_{2N_{n}-1}, S_{2N_{n-1}-1}, \ldots, S_1\}$
\STATE Memory matrices: $\{A_{1}, \ldots, A_{k}\}$
\FOR{each scan $S_i$ in $K$}
    \STATE $x'_{o}$ : $(T, B, E)$ $\leftarrow$ Conv1D$(x)$
    \STATE $B_o, C_o, \Delta_o$ $\leftarrow$ Linear$(x'_{o})$
    \STATE Transform $\overline{A}_{o}$ and $\overline{B}_{o}$ using $\Delta_o$
    \STATE $O_i$ $\leftarrow$ SSM$_{A_i,x}(x'_{o})$
\ENDFOR
\STATE \textcolor{gray}{\text{/* aggregation of outputs */}}
\STATE $z_{\text{HTM}}$ : $(T, B, E)$ $\leftarrow$ Linear(Aggregate$(\{O_{1}, \ldots, O_{k}\})$)
\STATE Return: $z_{\text{HTM}}$
\end{algorithmic}
\end{algorithm}

The encoder blocks are represented as $E_{1..N}$, arranged sequentially, and the decoder blocks as $D_{1..N}$, configured in reverse order to facilitate effective bottom-up and top-down information flow. Given that selective operations have  significantly lower computational complexity compared to attention-based methods, we have increased the number of scans to achieve higher quality generations. Concurrently, it is imperative to maintain a balance between the model's parameters and its efficiency. Thereby, a novel aspect of our model is the introduction of a hierarchical scan strategy, characterized by a sequence of scan numbers as,
\begin{equation}
K = \{S_{2N-1}, S_{2(N-1)-1}, \ldots, S_1\}.
\end{equation}
This sequence specifies the number of scans allocated to each layer, in descending order of complexity. 
For instance, the uppermost encoder layer, $E_1$, and the lowermost decoder layer, $D_N$, are allocated $S_{2N-1}$ scans, indicating the highest scanning complexity. Conversely, the lowest encoder layer, $E_N$, and the uppermost decoder layer, $D_1$, are assigned $S_1$ scans, reflecting the lowest level of scanning complexity. 
\begin{equation}
E_i(S) = 
\begin{cases} 
S_{2N-1} & \text{for } i=1 \\
S_{2(N-i)-1} & \text{for } i=2,\ldots,N-1 \\
S_1 & \text{for } i=N
\end{cases}
\end{equation}

\begin{equation}
D_j(S) = 
\begin{cases} 
S_{2N-1} & \text{for } j=N \\
S_{2(N-j)-1} & \text{for } j=N-1,\ldots,2 \\
S_1 & \text{for } j=1
\end{cases}
\end{equation}

This hierarchical scanning approach ensures that processing capabilities are evenly distributed throughout the encoder-decoder architecture., facilitating a detailed and nuanced analysis of temporal sequences. Within this structured framework, each denoiser is equipped with a specialized Hierarchical Temporal Mamba (HTM) block, which serves to augment the model's ability to process temporal information effectively. Additionally, the proposed Motion Mamba incorporates an attention-based mixer block denoted as $M$, strategically integrated to enhance conditional fusion.

\begin{algorithm}[t]
\caption{Bidirectional Spatial Mamba (BSM) Block.}
\label{algo:bsm}
\small
\begin{algorithmic}[1]
\REQUIRE{compressed latent representations $z$ : $(T, B, C)$}
\ENSURE{transformed representations $z_{\text{BSM}}$ : $(C, B, E)$}
\STATE \textcolor{gray}{\text{/* dimension rearrangement */}}
\STATE $z'$ : $(C, B, T)$ $\leftarrow$ Rearrange$(z)$
\STATE \textcolor{gray}{\text{/* linear projection after normalization */}}
\STATE $z'$ : $(C, B, T)$ $\leftarrow$ Norm$(z)$
\STATE $x, z$ : $(C, B, E)$ $\leftarrow$ Linear$(z')$
\FOR{$o$ in \{forward, backward\}}
    \STATE $x'_{o}$ : $(C, B, E)$ $\leftarrow$ Conv1D$(x)$
    \STATE $B_o, C_o, \Delta_o$ $\leftarrow$ Linear$(x'_{o})$
    \STATE Transform $\overline{A}_{o}$ and $\overline{B}_{o}$ using $\Delta_o$
    \STATE $y_o$ $\leftarrow$ SSM$(\overline{A}_{o}, \overline{B}_{o}, C_o)$
\ENDFOR
\STATE \textcolor{gray}{\text{/* gating and summing outputs */}}
\STATE $z_{\text{BSM}}$ : $(C, B, T)$ $\leftarrow$ GateAndSum$(y_{forward}, y_{backward}, z)$
\STATE $z'_{\text{BSM}}$ : $(T, B, C)$ $\leftarrow$ Rearrange$(z)$ 
\STATE Return: $z'_{\text{BSM}}$
\end{algorithmic}
\end{algorithm}

\noindent\textbf{Hierarchical Temporal Mamba (HTM)} 
block processes compressed latent representations, denoted as $z$, with the dimensions $(T, B, C)$, of which procedure shown in Algorithm \ref{algo:htm}.  Here, $T$ signifies the temporal dimension, as specified in the Variational AutoEncoder (VAE) framework. Initially, the input $z$ is subjected to a linear projection layer, producing transformed representations $x$ and $z$ with dimension $E$. 
Our analysis revealed an increased density of motion within the lower-level feature spaces. Consequently, we developed a hierarchical scanning methodology that is executed at various depths of the network. This approach not only accommodates the diverse motion densities, but also significantly reduces computational overhead.
This step utilizes a hierarchically structured set of scans, $K=\{S_{2N_{n}-1}, S_{2N_{n-1}-1}, \ldots, S_1\}$, in conjunction with a corresponding series of memory matrices $\{A_{1}, \ldots, A_{k}\}$. Each sub-SSM scan first applies a 1-D convolution to $x$, resulting in $x'_{o}$. $x'_{o}$ is then linearly projected to derive $B_o$, $C_o$, and $\Delta_o$. These projections $B_o$, $C_o$
use $\Delta_o$ to effect transformations in $\overline{A}_{o}$ and $\overline{B}_{o}$, respectively. After executing a sequence of SSM scans $\{SSM_{A_1,x}, SSM_{A_2,x}, \ldots, SSM_{A_k,x}\}$, a set of outputs $\{O_{1}, \ldots, O_{k}\}$ is compiled. This collection is subsequently aggregated via a linear projection to obtain the final output of the HTM block.

\subsubsection{Bidirectional Spatial Mamba (BSM)} 
block focuses on enhancing latent representation learning through a novel approach of dimension rearrangement and bidirectional scanning, of which the process is shown in Algorithm \ref{algo:bsm}. Initially, it alters the original input dimensions from $(T, B, C)$ to $(C, B, T)$, effectively swapping the temporal and channel dimensions. After this rearrangement, the input, now denoted $z'$, undergoes a linear projection after normalization, resulting in dimensions $x$ and $z$ of size $E$. The process involves bidirectional scanning of the latent channel dimension, where $\mathbf{x}$ is subjected to a 1-D convolution, yielding $\mathbf{x}'_{o}$ for both forward and backward directions. Each $\mathbf{x}'_{o}$ is then linearly projected to obtain $B_{o}$, $C_{o}$, and $\Delta_{o}$, which are utilized to transform $\overline{A}_{o}$ and $\overline{B}_{o}$, respectively. The final output token sequence, $\mathbf{z}_{\mathtt{l}}$, is computed by gating and summing the forward ${y}_{\text{forward}}$ and backward ${y}_{\text{backward}}$ output with $\mathbf{z}$.
This component is engineered to decode the structured latent skeleton by analyzing data from both forward and reverse viewpoints. Its main objective is to ensure the seamless continuity of information flow, thereby substantially enhancing the model's ability to generate accurate motion. This is achieved through the maintenance of a dense informational exchange, which is critical for the model's performance.

\section{Experiments}

\subsection{Datasets}

We evaluate our proposed Motion Mamba on two prominent \textit{Text-to-Motion} synthesis benchmarks as follows:

\noindent\textbf{HumanML3D.} 
The HumanML3D \cite{guo2022generating} dataset aggregates 14,616 motions sourced from the AMASS~\cite{mahmood2019amass} and HumanAct12~\cite{guo2020action2motion} datasets, with each motion accompanied by three textual descriptions, culminating in 44,970 scripts. This dataset spans a wide range of actions such as exercising, dancing, and acrobatics, presenting a rich motion-language corpus. 

\noindent\textbf{KIT-ML.} 
The KIT-ML dataset \cite{plappert2016kit} is comprised of 3,911 motions paired with 6,278 textual descriptions, serving as a compact yet effective benchmark for evaluation. For both datasets, the pose representation adopted is derived from T2M~\cite{guo2022generating}, ensuring consistency in motion representation across evaluations.

\subsection{Evaluation Metrics}
We adapt the standard evaluation metrics on following aspects throughout our experiments, including:
\noindent\textbf{Generation Quality.}
We implement a Fréchet inception distance (FID) \cite{heusel2017gans} to quantify the realism and diversity of motion generated by models. Moreover, we use multi-modal distance (MM Dist) to measure the distance between motions and texts and assess motion-text alignment.
\noindent\textbf{Diversity.}
We use the diversity metric to measure motion diversity, which calculates variance in features extracted from the motions. Additionally, we employ multi-modality (MModality) to assess diversity within generated motions sharing the same text description.

\begin{table}[t]
\caption{Comparison of text-conditional motion synthesis on HumanML3D~\cite{guo2022generating}. These metrics are evaluated by the motion encoder from \cite{guo2022generating}. Empty MModality indicates the non-diverse generation methods. We employ real motion as a reference and sort all methods by descending FIDs. The right arrow $\rightarrow$ means that the closer to the real motion, the better. \textbf{Bold} and \underline{underline} indicate the best and second best result.}
\vspace{-0.4cm}
\resizebox{\columnwidth}{!}{%
\begin{tabular}{@{}lccccccc@{}}
\toprule
\multirow{2}{*}{Method} & \multicolumn{3}{c}{R Precision $\uparrow$}                                                                                                                & \multicolumn{1}{c}{\multirow{2}{*}{FID$\downarrow$}} & \multirow{2}{*}{MM Dist$\downarrow$}              & \multirow{2}{*}{Diversity$\rightarrow$}           & \multirow{2}{*}{MModality$\uparrow$}              \\ \cmidrule(lr){2-4}
              & \multicolumn{1}{c}{Top 1} & \multicolumn{1}{c}{Top 2} & \multicolumn{1}{c}{Top 3} & \multicolumn{1}{c}{}                     &                          &                            &                            \\ \midrule
Real &
  $0.511^{\pm.003}$ &
  $0.703^{\pm.003}$ &
  $0.797^{\pm.002}$ &
  $0.002^{\pm.000}$ &
  $2.974^{\pm.008}$ &
  $9.503^{\pm.065}$ &
  \multicolumn{1}{c}{-}
  \\ \midrule
Seq2Seq \cite{plappert2018learning} &
  $0.180^{\pm.002}$ &
  $0.300^{\pm.002}$ &
  $0.396^{\pm.002}$ &
  $11.75^{\pm.035}$ &
  $5.529^{\pm.007}$ &
  $6.223^{\pm.061}$ &
  \multicolumn{1}{c}{-} \\
LJ2P \cite{ahuja2019language2pose}&
  $0.246^{\pm.001}$ &
  $0.387^{\pm.002}$ &
  $0.486^{\pm.002}$ &
  $11.02^{\pm.046}$ &
  $5.296^{\pm.008}$ &
  $7.676^{\pm.058}$ &
  \multicolumn{1}{c}{-} \\
T2G\cite{bhattacharya2021text2gestures} &
  $0.165^{\pm.001}$ &
  $0.267^{\pm.002}$ &
  $0.345^{\pm.002}$ &
  $7.664^{\pm.030}$ &
  $6.030^{\pm.008}$ &
  $6.409^{\pm.071}$ &
  \multicolumn{1}{c}{-} \\
Hier \cite{ghosh2021synthesis}&
    $0.301^{\pm.002}$ &
    $0.425^{\pm.002}$ &
    $0.552^{\pm.004}$ &
    $6.532^{\pm.024}$ &
    $5.012^{\pm.018}$ &
    $8.332^{\pm.042}$ &
    \multicolumn{1}{c}{-} \\
TEMOS \cite{petrovich22temos}&
  $0.424^{\pm.002}$ &
  $0.612^{\pm.002}$ &
  $0.722^{\pm.002}$ &
  $3.734^{\pm.028}$ &
  $3.703^{\pm.008}$ &
  $8.973^{\pm.071}$ &
  $0.368^{\pm.018}$ \\
T2M \cite{guo2022generating}&
  $0.457^{\pm.002}$ &
  $0.639^{\pm.003}$ &
  $0.740^{\pm.003}$ &
  $1.067^{\pm.002}$ &
  $3.340^{\pm.008}$ &
  $9.188^{\pm.002}$ &
  $2.090^{\pm.083}$ \\
MDM \cite{tevet2022human}&
  $0.320^{\pm.005}$ &
  $0.498^{\pm.004}$ &
  $0.611^{\pm.007}$ &
  ${0.544}^{\pm.044}$ &
  $5.566^{\pm.027}$ &
  $\textbf{9.559}^{\pm.086}$ &
  $\boldsymbol{2.799}^{\pm.072}$ \\
 MotionDiffuse \cite{zhang2024motiondiffuse} &
  $\underline{0.491}^{\pm.001}$ &
  $\underline{0.681}^{\pm.001}$ &
  $\underline{0.782}^{\pm.001}$ &
  $0.630^{\pm.001}$ &
  $\underline{3.113}^{\pm.001}$ &
  $\underline{9.410}^{\pm.049}$ &
  $1.553^{\pm.042}$ \\
MLD \cite{chen2023executing} &
  ${0.481}^{\pm.003}$ &
  ${0.673}^{\pm.003}$ &
  ${0.772}^{\pm.002}$ &
  $\underline{0.473}^{\pm.013}$ &
  ${3.196}^{\pm.010}$ &
  $9.724^{\pm.082}$ &
  $\underline{2.413}^{\pm.079}$ \\

 \midrule
\textbf{Motion Mamba (Ours)} &
    $\boldsymbol{0.502}^{\pm.003}$ &
    $\boldsymbol{0.693}^{\pm.002}$ &
    $\boldsymbol{0.792}^{\pm.002}$ &
    $\boldsymbol{0.281}^{\pm.009}$ &
    $\boldsymbol{3.060}^{\pm.058}$ &
    ${9.871}^{\pm.084}$ &
    $2.294^{\pm.058}$  

   \\ \bottomrule
\end{tabular}%
}
    \vspace{-0.2cm}
\label{tab:humanml3d}
\end{table}

\begin{table}[t]
\caption{We involve KIT-ML~\cite{plappert2016kit} dataset and evaluate the SOTA methods on the text-to-motion task.} 
\vspace{-0.4cm}
\resizebox{\columnwidth}{!}{%
\begin{tabular}{@{}lcccccccc@{}}
\toprule
\multirow{2}{*}{Method} & \multicolumn{3}{c}{R Precision $\uparrow$}                                                                                                                & \multicolumn{1}{c}{\multirow{2}{*}{FID$\downarrow$}} & \multirow{2}{*}{MM Dist$\downarrow$}              & \multirow{2}{*}{Diversity$\rightarrow$}           & \multirow{2}{*}{MModality$\uparrow$}              \\ \cmidrule(lr){2-4}
              & \multicolumn{1}{c}{Top 1} & \multicolumn{1}{c}{Top 2} & \multicolumn{1}{c}{Top 3} & \multicolumn{1}{c}{}                     &                          &                            &                            \\ \midrule
Real &
  $0.424^{\pm.005}$ &
  $0.649^{\pm.006}$ &
  $0.779^{\pm.006}$ &
  $0.031^{\pm.004}$ &
  $2.788^{\pm.012}$ &
  $11.08^{\pm.097}$ &
  \multicolumn{1}{c}{-}
  \\ \midrule
Seq2Seq\cite{plappert2018learning} &
  $0.103^{\pm.003}$ &
  $0.178^{\pm.005}$ &
  $0.241^{\pm.006}$ &
  $24.86^{\pm.348}$ &
  $7.960^{\pm.031}$ &
  $6.744^{\pm.106}$ &
  \multicolumn{1}{c}{-} \\
T2G\cite{bhattacharya2021text2gestures} &
  $0.156^{\pm.004}$ &
  $0.255^{\pm.004}$ &
  $0.338^{\pm.005}$ &
  $12.12^{\pm.183}$ &
  $6.964^{\pm.029}$ &
  $9.334^{\pm.079}$ &
  \multicolumn{1}{c}{-} \\
LJ2P \cite{ahuja2019language2pose}&
  $0.221^{\pm.005}$ &
  $0.373^{\pm.004}$ &
  $0.483^{\pm.005}$ &
  $6.545^{\pm.072}$ &
  $5.147^{\pm.030}$ &
  $9.073^{\pm.100}$ &
  \multicolumn{1}{c}{-} \\
Hier \cite{ghosh2021synthesis}&
  $0.255^{\pm.006}$ &
  $0.432^{\pm.007}$ &
  $0.531^{\pm.007}$ &
  $5.203^{\pm.107}$ &
  $4.986^{\pm.027}$ &
  $9.563^{\pm.072}$ &
  $\underline{2.090}^{\pm.083}$ \\
TEMOS \cite{petrovich22temos}&
    $0.353^{\pm.006}$ & 
    $0.561^{\pm.007}$ & 
    $0.687^{\pm.005}$ & 
    $3.717^{\pm.051}$ & 
    $3.417^{\pm.019}$ & 
    $10.84^{\pm.100}$ & 
    $0.532^{\pm.034}$ \\
T2M \cite{guo2022generating}&
  $0.370^{\pm.005}$ &
  $0.569^{\pm.007}$ &
  $0.693^{\pm.007}$ &
  $2.770^{\pm.109}$ &
  $3.401^{\pm.008}$ &
  ${10.91}^{\pm.119}$ &
  $1.482^{\pm.065}$ \\
MDM \cite{tevet2022human}&
  $0.164^{\pm.004}$ &
  $0.291^{\pm.004}$ &
  $0.396^{\pm.004}$ &
  ${0.497}^{\pm.021}$ &
  $9.191^{\pm.022}$ &
  $10.85^{\pm.109}$ &
  ${1.907}^{\pm.214}$ \\
MotionDiffuse \cite{zhang2024motiondiffuse} &
$\underline{0.417}^{\pm.004}$ &
$\underline{0.621}^{\pm.004}$ &
$\underline{0.739}^{\pm.004}$ &
$1.954^{\pm.062}$ &
$\textbf{2.958}^{\pm.005}$ &
$\textbf{11.10}^{\pm.143}$ &
$0.730^{\pm.013}$ \\
MLD \cite{chen2023executing} &
${0.390}^{\pm.008}$ & 
${0.609}^{\pm.008}$ & 
${0.734}^{\pm.007}$ & 
$\underline{0.404}^{\pm.027}$ & 
${3.204}^{\pm.027}$ & 
$10.80^{\pm.117}$ & 
$\textbf{2.192}^{\pm.071}$ \\
\midrule
\textbf{Motion Mamba (Ours)} &
    $\boldsymbol{0.419}^{\pm.006}$ &
    $\boldsymbol{0.645}^{\pm.005}$ &
    $\boldsymbol{0.765}^{\pm.006}$ &
    $\boldsymbol{0.307}^{\pm.041}$ &
    $\underline{3.021}^{\pm.025}$ &
    $\underline{11.02}^{\pm.098}$ &
    $1.678^{\pm.064}$
   \\ \bottomrule
\end{tabular}%
}
\label{tab:kit}
    \vspace{-0.4cm}
\end{table}

\subsection{Comparative Studies}
We evaluate our method against the state-of-the-art methods on the HumanML3D \cite{guo2022generating} and KIT-ML \cite{plappert2016kit} datasets.
We train our Motion Mamba with HTM arrangement strategy MM ($\{S_{2N_{n}-1}, \ldots, S_1\}$), BSM bidirectional block strategy on the latent dimension = 2 with 11 layers. We evaluate our model and previous works with suggested metrics in HumanML3D \cite{guo2022generating} and calculate 95\% confidence interval by repeat evaluation 20 times. 
The results for the HumanML3D dataset are presented in Table \ref{tab:humanml3d}. Our model outperforms other methods significantly across various evaluation metrics, including FID, R precision, multi-modal distance, and diversity. For instance, our Motion Mamba outperforms previous best diffusion based motion generation MLD by 40.5\% in terms of FID, and up to 10\% improvement on R Precision, we aslo obatined best MModality by 3.060. The results for the KIT-ML dataset are presented in Table \ref{tab:kit}. We have also outperformed other well-established methods in FID and multi-modal distance.

\begin{table}[t]
\caption{In order to evaluate the models' capability in long sequence motion generation, we compared our method with an existing approach on the recently introduced HumanML3D-LS dataset. This dataset comprises motion sequences longer than 190 frames from the original evaluation set. Our model demonstrates superior performance compared to other methods.} 
\vspace{-0.4cm}
\resizebox{\columnwidth}{!}{%
\begin{tabular}{@{}lcccccccc@{}}
\toprule
\multirow{2}{*}{Method} & \multicolumn{3}{c}{R Precision $\uparrow$}                                                                                                                & \multicolumn{1}{c}{\multirow{2}{*}{FID$\downarrow$}} & \multirow{2}{*}{MM Dist$\downarrow$}              & \multirow{2}{*}{Diversity$\rightarrow$}           & \multirow{2}{*}{MModality$\uparrow$}              \\ \cmidrule(lr){2-4}
              & \multicolumn{1}{c}{Top 1} & \multicolumn{1}{c}{Top 2} & \multicolumn{1}{c}{Top 3} & \multicolumn{1}{c}{}                     &                          &                            &                            \\ \midrule
Real &
  $0.437^{\pm.003}$ &
  $0.622^{\pm.004}$ &
  $0.721^{\pm.004}$ &
  $0.004^{\pm.000}$ &
  $3.343^{\pm.015}$ &
  $8.423^{\pm.090}$ &
  \multicolumn{1}{c}{-}
  \\ \midrule
MDM \cite{tevet2022human}&
  $0.368^{\pm.005}$ &
  $0.553^{\pm.006}$ &
  $0.672^{\pm.005}$ &
  $\underline{0.802}^{\pm.044}$ &
  $3.860^{\pm.025}$ &
  $\underline{8.817}^{\pm.068}$ &
   -  \\
MotionDiffuse \cite{zhang2024motiondiffuse} &
${0.367}^{\pm.004}$ &
${0.521}^{\pm.004}$ &
${0.623}^{\pm.004}$ &
${2.460}^{\pm.062}$ &
${3.789}^{\pm.005}$ &
$\textbf{8.707}^{\pm.143}$ &
$1.602^{\pm.013}$ \\
MLD \cite{chen2023executing} &
$\underline{0.403}^{\pm.005}$ & 
$\underline{0.584}^{\pm.005}$ & 
$\underline{0.690}^{\pm.005}$ & 
${0.952}^{\pm.020}$ & 
$\underline{3.580}^{\pm.016}$ & 
${9.050}^{\pm.085}$ & 
$\textbf{2.711}^{\pm.104}$ \\
\midrule
\textbf{Motion Mamba (Ours)} &
    $\textbf{0.417}^{\pm.003}$ &
    $\textbf{0.606}^{\pm.003}$ &
    $\textbf{0.713}^{\pm.004}$ &
    $\textbf{0.668}^{\pm.019}$ &
    $\textbf{3.435}^{\pm.015}$ &
    ${9.021}^{\pm.070}$ &
    $\underline{2.373}^{\pm.084}$
   \\ \bottomrule
\end{tabular}%
}
    \vspace{-0.4cm}
\label{tab:long}
\end{table}

\subsection{Ablation Studies}

We concluded the ablation studies including long sequence evaluation, hierarchical design with HTM, bidirectional design in the BSM, number of latent dimensions, and number of layers of our proposed motion mamba in Table~\ref{tab:ablation:mm}. 

\begin{wrapfigure}{r}{0.4\textwidth} %
  \centering
      \vspace{-0.1cm}
    \includegraphics[width=1\linewidth]{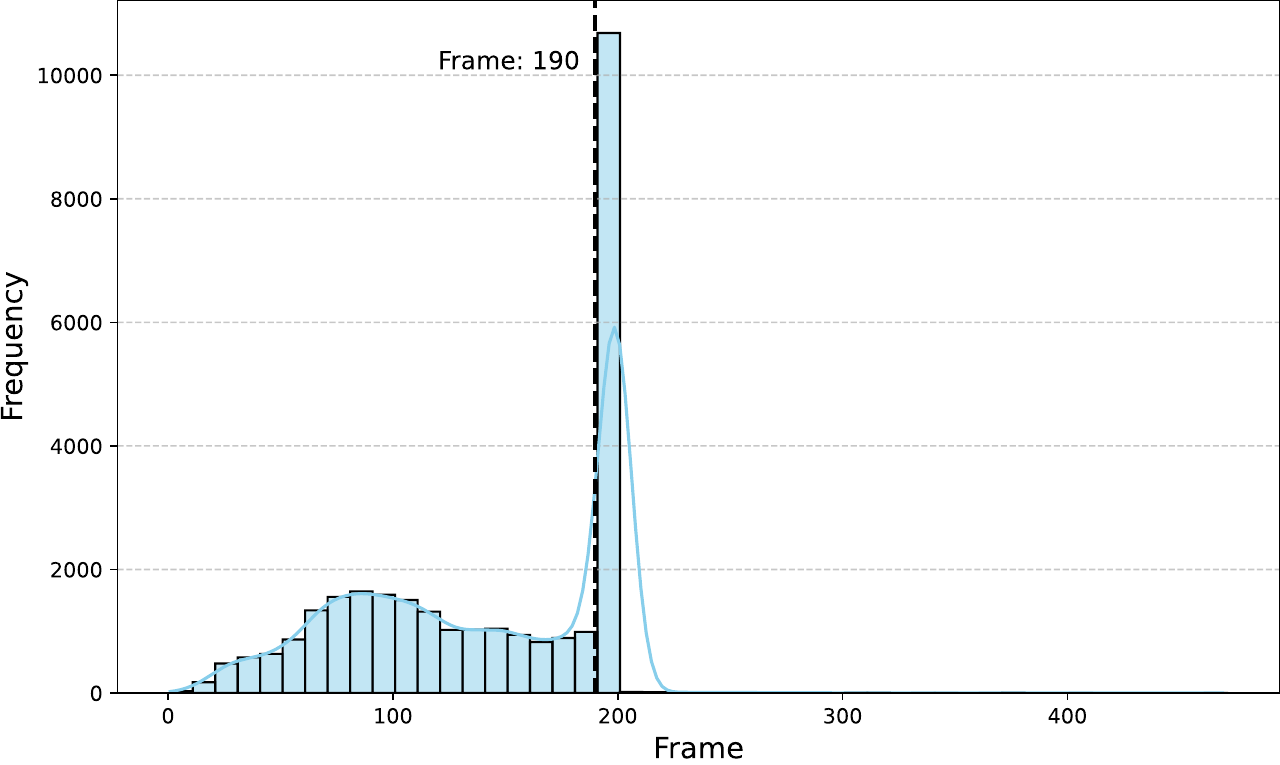} 
    \caption{The figure shows a long tail distribution of the HumanML3D \cite{guo2022generating}, which has a significant proportion of long-sequence human motions.}
        \label{fig:plot}
    \vspace{-0.8cm}
\end{wrapfigure}

\noindent\textbf{Long Sequence Motion Generation}. The HumanML3D \cite{guo2022generating} dataset exhibits a long-tailed and right-skewed distribution with a significant proportion of long-sequence human motions, as shown in Figure \ref{fig:plot}. We suggest previous studies overlooked the challenges in the long-sequence generation problem.
Thus, we introduce a new dataset variant, \textit{HumanML3D-LS}, comprising motion sequences longer than 190 frames extracted from the original test set. This addition allows us to showcase our capability in generating long-sequence motions.
Subsequently, we evaluate the performance of our method on HumanML3D-LS and compare it with other diffusion-based motion generation approaches. The comparative results are presented in Table~\ref{tab:long}. Motion Mamba by leverage the benefits on long-range dependency modeling make it well suitable for long sequence motion generation.

\noindent\textbf{Hierarchical Design with HTM.}
In our ablation studies, we observed a slight improvement upon reversing the scan order from a lower to a higher level, specifically transitioning from MM \(\{S_1, \ldots, S_N\}\) to MM \(\{S_N, \ldots, S_1\}\). This enhancement suggests a correlation with the increase in temporal motion density within the lower-level feature spaces. 
Furthermore, to achieve the optimal result, we introduce the hierarchical design to arrange the scanning frequency, resulting in the sequence MM \(\{S_{2N_n-1}, \ldots, S_1\}\). This expansion in the number of scans led to a performance increase. We attribute this enhancement to the observation that individual selective scan operations significantly reduce the parameter count, especially when compared to the parameter-intensive constructs of self-attention and feedforward network blocks prevalent in transformer architectures.

\begin{table}[t]
\centering
\caption{Evaluation of text-based motion synthesis on HumanML3D~\cite{guo2022generating}: we use metrics in Table \ref{tab:humanml3d} and provides real reference, we evaluate the various HTM and BSM design choices, the dimension of the latent input, the different number of layer of Motion Mamba model. 
}
\vspace{-0.4cm}
\resizebox{0.8\columnwidth}{!}{%
\begin{tabular}{@{}lccccc@{}}
\toprule
\multirow{2}{*}{Models} & \multicolumn{1}{c}{R Precision} & \multicolumn{1}{c}{\multirow{2}{*}{FID$\downarrow$}} & \multirow{2}{*}{MM Dist.$\downarrow$} & \multirow{2}{*}{Diversity$\rightarrow$} & \multirow{2}{*}{MModality$\uparrow$} \\
                           & \multicolumn{1}{c}{Top 3$\uparrow$}       & \multicolumn{1}{c}{}                     &                          &                            &     \\ \toprule
Real &
  $0.797^{\pm.002}$ &
  $0.002^{\pm.000}$ &
  $2.974^{\pm.008}$ &
  $9.503^{\pm.065}$ &
  \multicolumn{1}{c}{-}
  \\ \midrule
MM ($\{S_1, \ldots, S_{N}\}$) &
    $0.673^{\pm.003}$ &
    $1.278^{\pm.012}$ &
    $3.802^{\pm.041}$ &
    $8.678^{\pm.096}$ &
    $3.127^{\pm.024}$ \\
MM ($\{S_{N}, \ldots, S_1\}$) &
    $0.738^{\pm.002}$ &
    $0.962^{\pm.011}$ &
    $3.433^{\pm.003}$ &
    $9.180^{\pm.071}$ &
    $2.723^{\pm.033}$ \\
MM ($\{S_1, \ldots, S_{2N_{n-1}}\}$) &
    $0.698^{\pm.002}$ &
    $0.856^{\pm.008}$ &
    $3.624^{\pm.037}$ &
    $9.229^{\pm.067}$ &
    $2.826^{\pm.017}$ \\
\textbf{MM ($\{S_{2N_{n}-1}, \ldots, S_1\}$)}  &
	 $0.792^{\pm.002}$ &
    $0.281^{\pm.009}$ &
    $3.060^{\pm.058}$ &
    $9.871^{\pm.084}$ &
    $2.294^{\pm.058}$   \\
\midrule
MM ($Single Scan$) &
    $0.736^{\pm.003}$ &
    $1.063^{\pm.010}$ &
    $3.443^{\pm.026}$ &
    $9.180^{\pm.067}$ &
    $2.676^{\pm.041}$
\\
MM ($BiScan, layer$) &
    $0.735^{\pm.004}$ &
    $0.789^{\pm.007}$ &
    $3.408^{\pm.034}$ &
    $9.374^{\pm.059}$ &
    $2.591^{\pm.046}$
\\
\textbf{MM ($BiScan, block$) }&
	 $0.792^{\pm.002}$ &
    $0.281^{\pm.009}$ &
    $3.060^{\pm.058}$ &
    $9.871^{\pm.084}$ &
    $2.294^{\pm.058}$   \\
\midrule
MM ($Dim,1$) &
    $0.706^{\pm.003}$ &
    $0.652^{\pm.011}$ &
    $3.541^{\pm.072}$ &
    $9.141^{\pm.082}$ &
    $2.612^{\pm.055}$
\\
\textbf{MM ($Dim,2$)} &
	 $0.792^{\pm.002}$ &
    $0.281^{\pm.009}$ &
    $3.060^{\pm.058}$ &
    $9.871^{\pm.084}$ &
    $2.294^{\pm.058}$   \\
MM ($Dim,5$) &
    $0.741^{\pm.008}$ &
    $0.728^{\pm.009}$ &
    $3.307^{\pm.027}$ &
    $9.427^{\pm.099}$ &
    $2.314^{\pm.062}$
\\
MM ($Dim,7$) & 
    $0.738^{\pm.004}$ &
    $0.599^{\pm.007}$ &
    $3.359^{\pm.068}$ &
    $9.166^{\pm.075}$ &
    $2.488^{\pm.037}$ 
\\
MM ($Dim,10$) &
    $0.715^{\pm.003}$ &
    $0.628^{\pm.008}$ &
    $3.548^{\pm.043}$ &
    $9.200^{\pm.075}$ &
    $2.884^{\pm.096}$
\\
\midrule
MM (9 layers) &
    $0.755^{\pm.002}$ &
    $1.080^{\pm.012}$ &
    $3.309^{\pm.057}$ &
    $9.721^{\pm.081}$ &
    $2.974^{\pm.039}$ 
\\
\textbf{MM (11 layers)} &
	 $0.792^{\pm.002}$ &
    $0.281^{\pm.009}$ &
    $3.060^{\pm.058}$ &
    $9.871^{\pm.084}$ &
    $2.294^{\pm.058}$   \\
MM (27 layers) & 
    $0.750^{\pm.003}$ &
    $0.975^{\pm.008}$ &
    $3.336^{\pm.096}$ &
    $9.249^{\pm.071}$ &
    $2.821^{\pm.063}$
\\
MM (37 layers) &
    $0.754^{\pm.005}$ &
    $0.809^{\pm.010}$ &
    $3.338^{\pm.061}$ &
    $9.355^{\pm.062}$ &
    $2.741^{\pm.077}$
\\
\bottomrule
\end{tabular}%
}
\vspace{-0.4cm}
\label{tab:ablation:mm}
\end{table}

\noindent\textbf{Bidirectional Design in BSM.}
We developed three distinct variations of latent scanning mechanisms, differentiated by their scanning directions. In the context of motion generation tasks, we posit that the flow of hidden information within the structured latent skeleton holds significance, an aspect previously underexplored. Our ablation study reveals that a $single~scan$ across the latent dimension yields minimal improvement. Subsequently, we investigated both layer-based and block-based bidirectional scans. Our findings indicate that the block-based bidirectional scan achieves optimal performance. This suggests that spatial information flows are encoded within the latent spaces and that bidirectional scanning facilitates the exchange of this information, thereby enhancing the efficacy of motion generation tasks.

\noindent\textbf{Architecture Design for Motion Mamba.}
The proposed Motion Mamba which is grounded in a standardized motion latent diffusion system. We delved into the interplay between dimensional aspects and the module's capacity (measured by the number of layers) to ascertain their impact on system performance. Experimental results demonstrate that the Motion Mamba achieves superior performance at a latent dimension of 2, diverging from prior works where the optimal dimension was identified as 1. We attribute this discrepancy to our HTM, which necessitates multiple scans correlating with the sequence length, thus implicating dimensionality as a pivotal factor. A marginal increase in dimensionality enabled us to attain peak performance, simultaneously enhancing efficiency compared to models with a dimensionality of 10.
Furthermore, we conducted experiments to determine the optimal layer count for Motion Mamba, inspired by the design of its selective scanning mechanism. Notably, a single Mamba layer achieves a parameter reduction of approximately 75\% compared to a conventional transformer encoder block. By increasing the number of layers, we aim to uncover the relationship between model capacity and its performance. Our findings reveal that, through the integration of our specially designed HTM and BSM (Bidirectional Scanning Module) blocks, the Motion Mamba reaches its optimal performance with 11 layers. This represents a slight increase over the MLD \cite{chen2023executing} baseline. However, due to the reduced parameter count in each layer, Motion Mamba exhibits significantly greater efficiency than previous methodologies.

\begin{wrapfigure}{r}{0.45\textwidth} %
  \centering
      \vspace{-1.4cm}
    \includegraphics[width=1\linewidth]{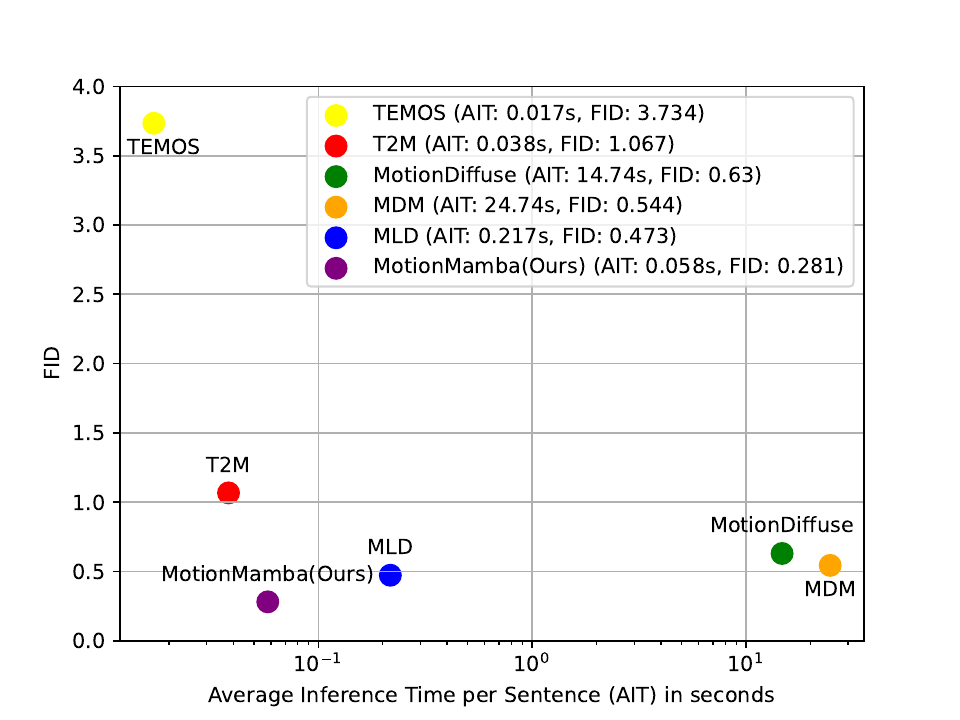} 
    \caption{The figure shows the average inference time per sentence (AIT) vs FID, our proposed motion mamba obtained 0.058s AIT and 0.281 FID overall outperform previous methods. We evaluate all methods on a single V100 GPU.}
        \label{fig:inference}
    \vspace{-1.2cm}
\end{wrapfigure}

\subsection{Inference Time}
Inference time remains a significant challenge for diffusion-based methods. To address this, we enhance the inference speed by incorporating the efficient Mamba block within a lightweight architecture. Compared to the previous strong baseline, such as the MLD model cited in \cite{chen2023executing}, which reports an average inference time of 0.217 seconds, our Motion Mamba model achieves a notable reduction in computational overhead, as shown in Figure \ref{fig:inference}. Specifically, it requires four times less computational effort, thereby facilitating faster and real-time inference speeds.

\section{Discussion and Conclusion}

In this study, we introduced Motion Mamba, a novel framework designed for efficient and extended sequence motion generation. Our approach represents the inaugural integration of the Mamba model within the domain of motion generation, featuring significant advancements including the implementation of Hierarchical Temporal Mamba (HTM) blocks. These blocks are specifically engineered to enhance temporal alignment through hierarchically organized selective scanning. Furthermore, Bidirectional Spatial Mamba (BSM) blocks have been developed to amplify the exchange of information flow within latent spaces, thereby augmenting the model's ability to bidirectionally capture skeleton-level density features with greater precision.
Compared to previous diffusion-based motion generation methodologies that predominantly utilize transformer blocks, our Motion Mamba framework achieves SOTA performance, evidencing an improvement of up to 50\% in FID scores and a quadrupled improvement in inference speed. Through comprehensive experimentation across a variety of human motion generation tasks, the effectiveness and efficiency of our proposed Motion Mamba model have been robustly demonstrated, marking a significant leap forward in the field of human motion generation.

\section*{Acknowledgements}
This work is partially supported by the Fundamental Research Funds for the Central Universities, Peking University.

\clearpage

\title{
 Motion Mamba Supplementary
} 

\titlerunning{Motion Mamba Supplementary}

\authorrunning{Zeyu and Akide et al.}

\author{}

\institute{}

\maketitle

\begin{figure}[b!]
\centering

\begin{subfigure}{.5\textwidth}
  \centering
  \includegraphics[width=\linewidth]{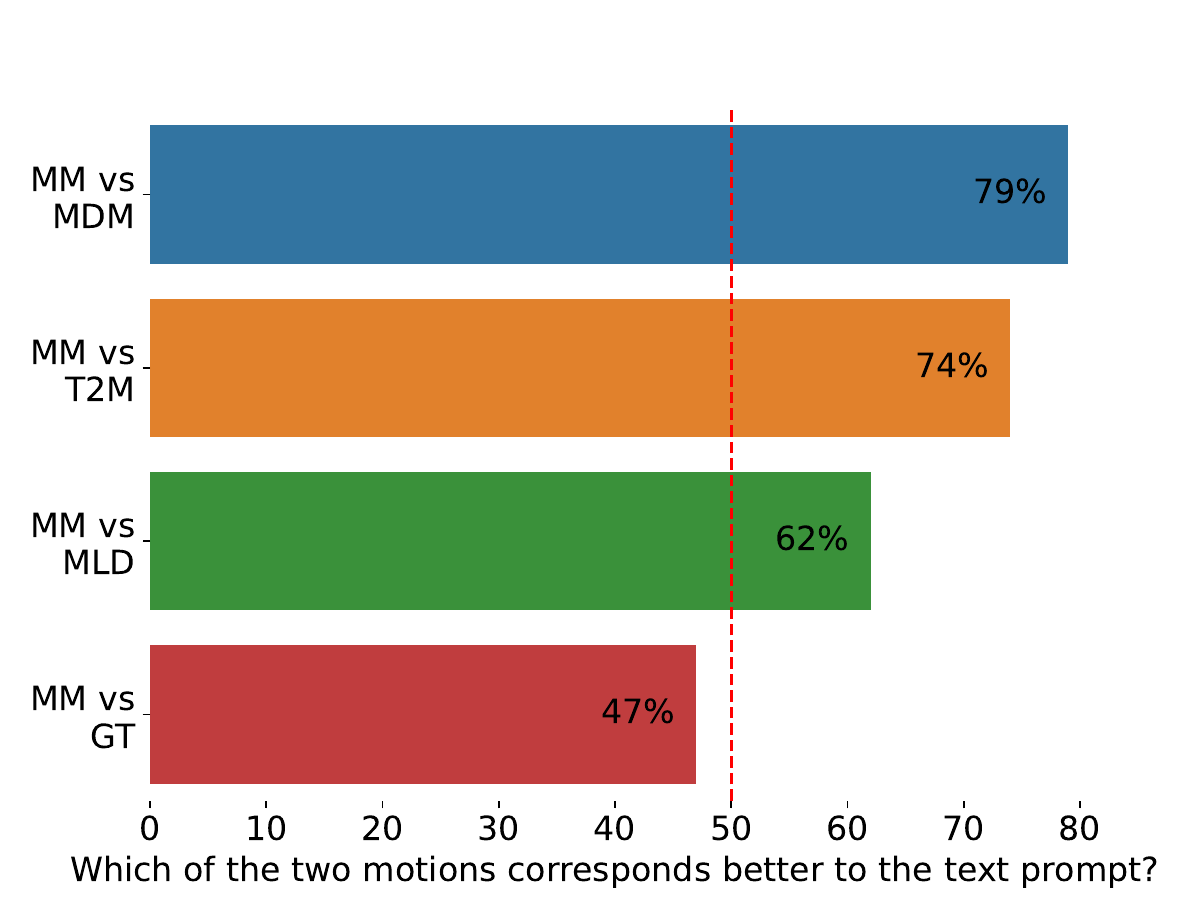}
  \caption{Text-Motion Correspondence User Study}
  \label{fig:us_sub1}
\end{subfigure}%
\begin{subfigure}{.5\textwidth}
  \centering
  \includegraphics[width=\linewidth]{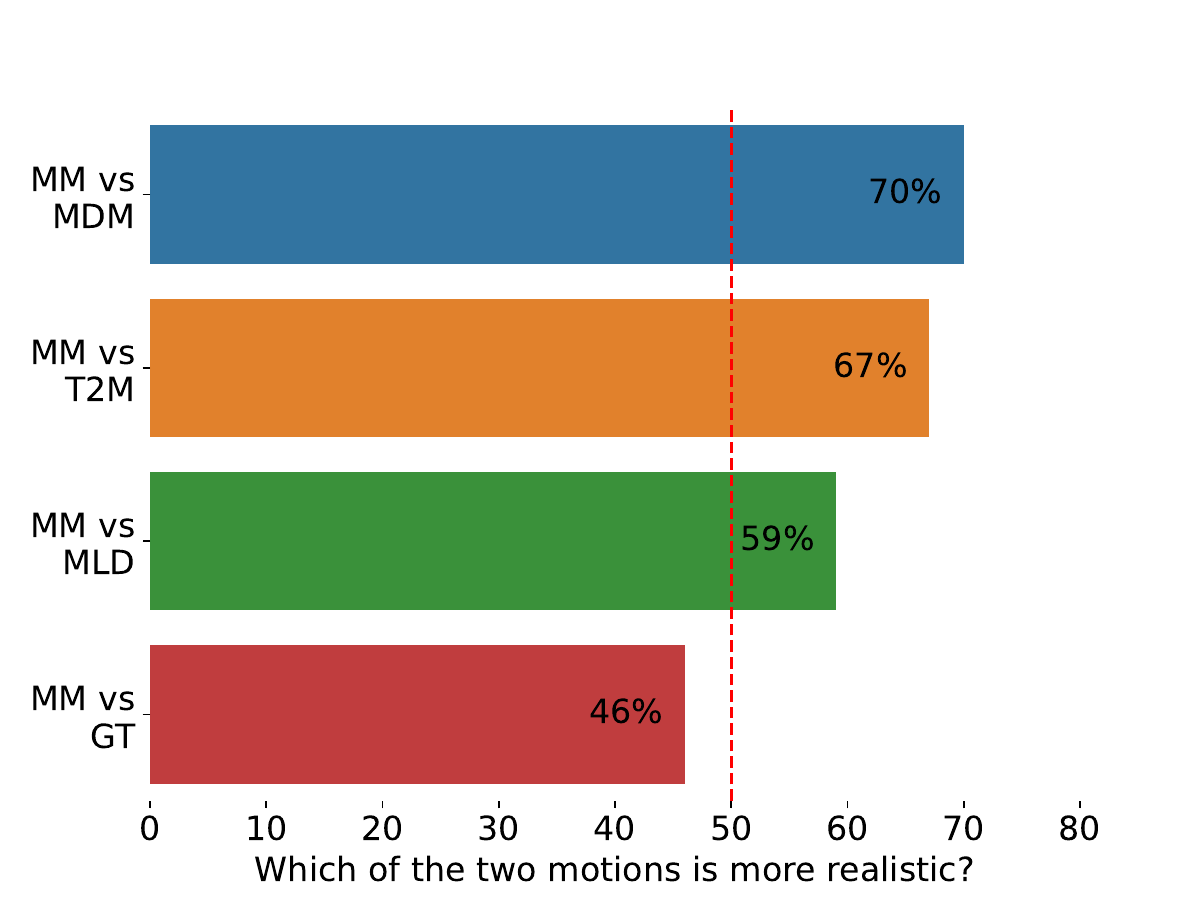}
  \caption{Quality User Study}
  \label{fig:us_sub2}
\end{subfigure}

\caption{User Study in two aspects including text-motion correspondence and quality, we compare \textit{Motion Mamba}(MM) with previous methods including MDM \cite{tevet2022human} , T2M \cite{guo2022generating} , MLD \cite{chen2023executing} and ground truth.}
\label{fig:us}

\end{figure}

\section{Implementation Details}

Motion Mamba operates within the latent spaces, leveraging the capabilities of the Motion Variational AutoEncoder (VAE) $\mathcal{V}=\{\mathcal{E}, \mathcal{D}\}$, as proposed in the seminal work by Chen et al. \cite{chen2023executing}. For the configuration of the Motion Mamba denoiser $\epsilon_\theta$, we have opted for an architecture comprising 11 layers ($N = 11$), with the latent dimensionality set to $z \in \mathbb{R}^{2,d}$. The Hierarchical Temporal Mamab (HTM) modules are arranged in a scan pattern of $\{S^{2N_{n}-1}, \ldots, S^1\}$, while the Bidirectional Spatial (BSH) modules incorporate a block-level bidirectional scan policy. Additionally, we utilize a pretrained \textit{CLIP-VIT-L-14} model in a frozen state to derive text embeddings $\tau_\theta^w(w^{1:N}) \in \mathbb{R}^{1 \times d}$. 

All models under the Motion Mamba framework are meticulously trained using the AdamW Optimizer, with the learning rate steadfastly maintained at $10^{-4}$. We have standardized our global batch size at 512, which is judiciously distributed across 4 GPUs to facilitate data-parallel training. The training regime is extended over 2,000 epochs to ensure convergence to an optimal set of parameters. For the diffusion sampling process, we maintain the number of steps at 1,000 and 50 during the training and inference phases, respectively. The entire training procedure is executed on a single-node GPU server, outfitted with 4 NVIDIA A100 GPUs, spanning approximately 4 hours. Inference speed evaluations of our Motion Mamba models are conducted on a single NVIDIA V100 GPU for fair comparison, while module development and additional inference tasks are performed on a single NVIDIA GeForce RTX 3090/4090 GPU.

\section{User Study}

In this work, we undertake a comprehensive evaluation of \textit{Motion Mamba}'s performance, encompassing both qualitative analyses across various datasets and a user study to assess its real-world applicability. A diverse collection of 20 motion sequence sets, prompted randomly and extracted from the HumanML3D \cite{guo2022generating} test set, were generated utilizing three distinct methodologies—MDM \cite{tevet2022human}, T2M \cite{guo2022generating}, MLD \cite{chen2023executing}—alongside \textit{Motion Mamba} and a baseline of ground truth motions. Subsequently, 50 participants were randomly selected to evaluate the motion sequences generated by these methods.

\begin{figure}[t]
    \centering
    \includegraphics[width=0.9\linewidth]{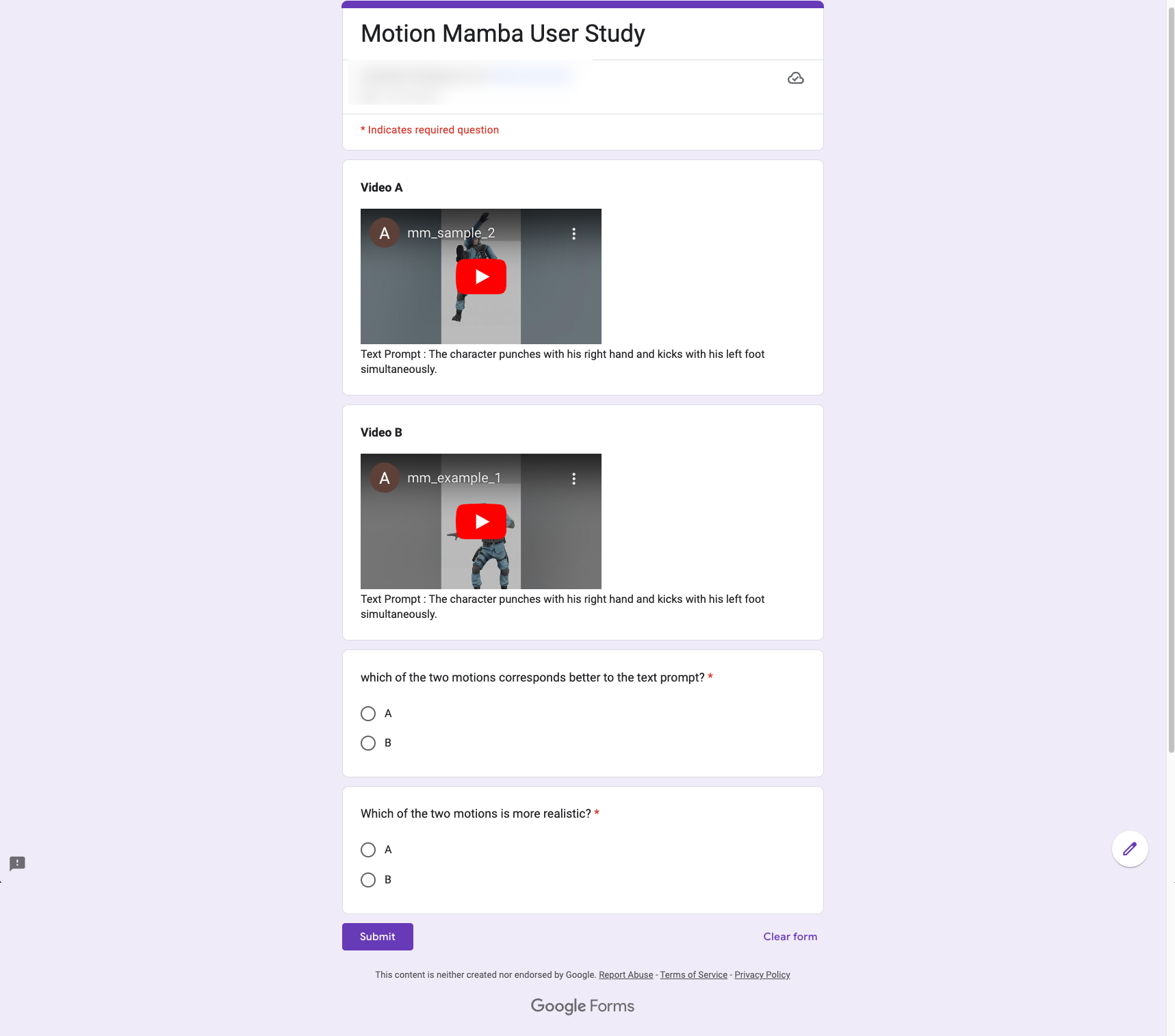} 
    \caption{
    This figure presents the User Interface (UI) deployed for our User Study, wherein participants are presented with two videos, labeled as Video A and Video B, respectively. These videos are selected randomly from a pool consisting of outputs generated by three distinct methods, in addition to the Ground Truth (GT) for comparison. Participants are posed with two types of evaluative questions to gauge the effectiveness of the generated motions. The first question, "Which of the two motions is more realistic?", aims to assess the overall quality and realism of the motion capture. The second question, "Which of the two motions corresponds more accurately to the text prompt?", is designed to evaluate the congruence between the generated motion and the provided text prompt. This dual-question approach facilitates a comprehensive assessment of both the quality of the motion generation and its fidelity to the specified text prompts.
    }
    \label{fig:ui}
    \vspace{-0.4cm}
\end{figure}

The user study was administered through a Google Forms interface, as depicted in \cref{fig:ui}, ensuring that motion sequences were presented anonymously without revealing their generative model origins. Our analysis focused on two critical dimensions: the fidelity of text-to-motion correspondence and the overall quality of the generated motions.

Empirical results, illustrated in \cref{fig:us_sub1} and \cref{fig:us_sub2}, unequivocally demonstrate \textit{Motion Mamba}'s superior performance relative to the benchmark methods in terms of both text-motion alignment and motion quality. Specifically, \textit{Motion Mamba} achieved significant margins over MDM \cite{tevet2022human}, T2M \cite{guo2022generating}, and MLD \cite{chen2023executing} by 79\%, 74\%, and 62\% in text-motion correspondence, respectively, as highlighted in \cref{fig:us_sub1}. When juxtaposed with ground truth data—meticulously captured with state-of-the-art, noise-free devices—\textit{Motion Mamba}'s generated sequences exhibited a remarkably close adherence to the intended text descriptions, underscoring its proficiency in aligning textual prompts with motion sequences.

Further reinforcing these findings, \textit{Motion Mamba}'s generated motions were also found to surpass the aforementioned methods by substantial margins of 70\%, 67\%, and 59\%, respectively, in terms of quality, as reported in \cref{fig:us_sub2} . This underscores \textit{Motion Mamba}'s ability to not only closely match the text-motion correspondence of high-fidelity ground truth data but also to produce high-quality motion sequences that resonate well with real user experiences.

\section{Visualization}

Our study delves into the visualization of motion generation by capturing intricate motion sequences, utilizing prompts and their variations derived from HumanML3D \cite{guo2022generating}. We meticulously compare our proposed Motion Mamba methodology with established state-of-the-art techniques, namely MotionDiffuse \cite{zhang2024motiondiffuse}, MDM \cite{tevet2022human}, and MLD \cite{chen2023executing}. Presenting three distinct motion sequences, we meticulously analyze and visualize each, offering a comprehensive assessment of our approach's efficacy.

\begin{figure}[!t]
    \centering
    \includegraphics[width=1.0\linewidth]{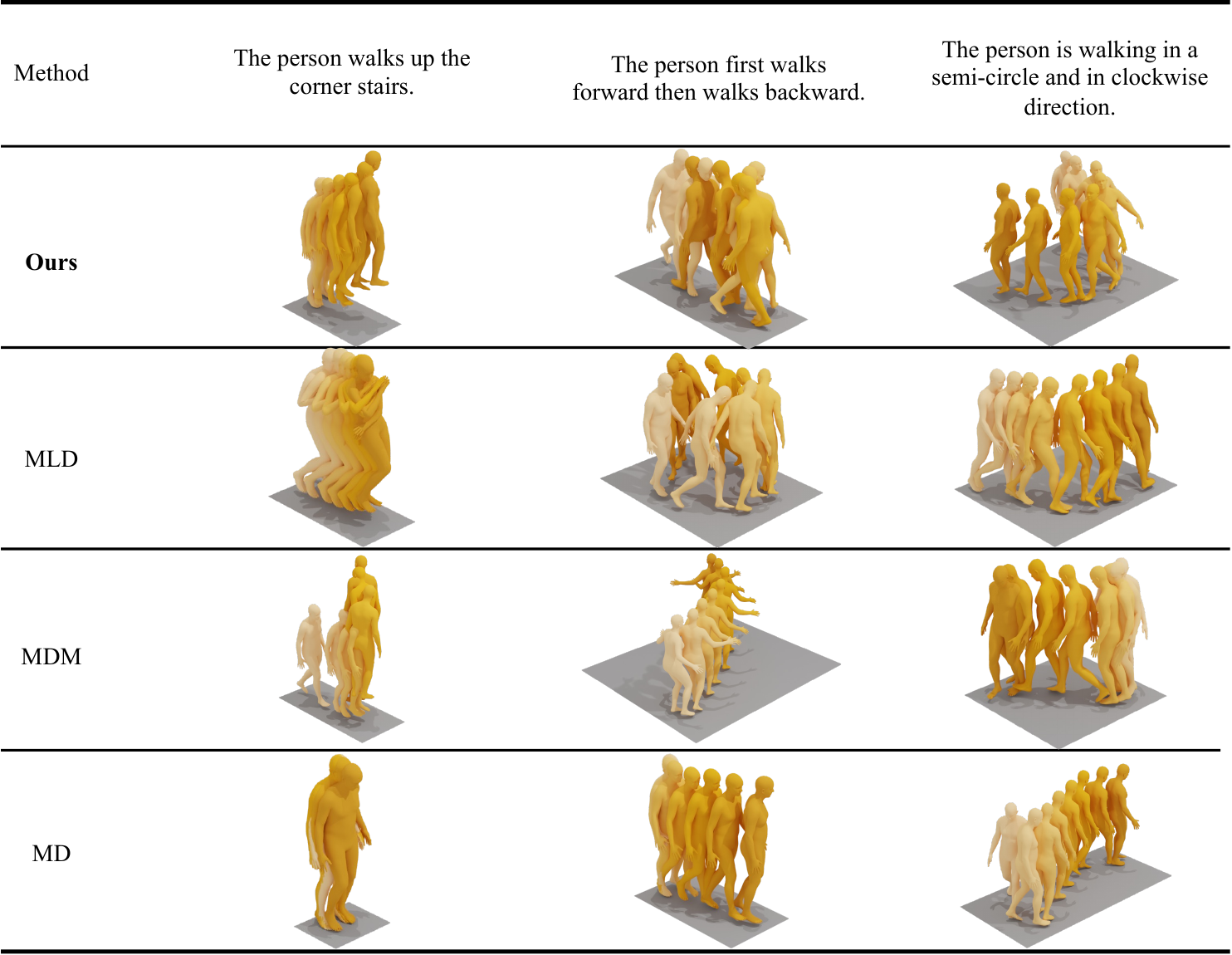} 
    \caption{We compared the proposed Motion Mamba with well-established state-of-the-art methods such as MotionDiffuse \cite{zhang2024motiondiffuse}, MDM \cite{tevet2022human}, and MLD \cite{chen2023executing}. We presented three distinct motion prompts and visualized them in the form of motion sequence. The results demonstrated our superior performance compared to existing methods.}
    \label{fig:compare}
    \vspace{-0.4cm}
\end{figure}

\begin{figure}[t]
    \centering
    \includegraphics[width=0.9\linewidth]{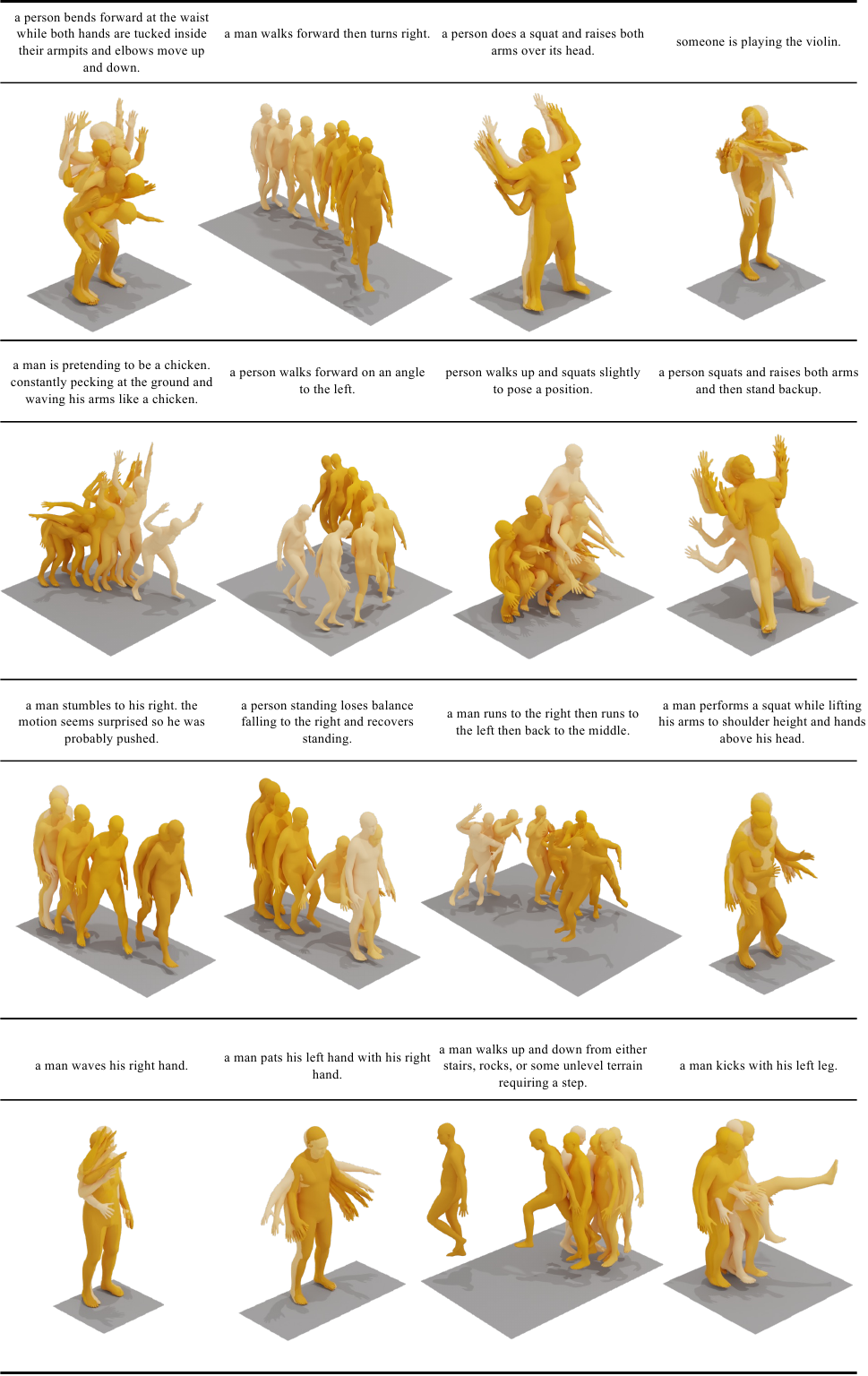} 
    \caption{We have included extra examples to showcase the proposed Motion Mamba model. These examples feature randomly selected prompts sourced from HumanML3D \cite{guo2022generating}, providing additional visualizations of the model's capabilities.}
    \label{fig:visual}
    \vspace{-0.4cm}
\end{figure}

% \clearpage  
% \bibliographystyle{splncs04}
% \bibliography{main}

\end{document}